\documentclass{article}

\PassOptionsToPackage{numbers,sort&compress}{natbib}
\usepackage[preprint]{neurips_2026}
\usepackage{amsmath}
\usepackage{comment}

\usepackage[utf8]{inputenc} % allow utf-8 input
\usepackage[T1]{fontenc}    % use 8-bit T1 fonts
\usepackage{hyperref}       % hyperlinks
\usepackage{url}            % simple URL typesetting
\usepackage{booktabs}       % professional-quality tables
\usepackage{amsfonts}       % blackboard math symbols
\usepackage{nicefrac}       % compact symbols for 1/2, etc.
\usepackage{microtype}      % microtypography
\usepackage[table]{xcolor}         % colors
\usepackage{graphicx}
\usepackage{subcaption}
\usepackage{xspace}
\usepackage{fancyhdr}
\hypersetup{hidelinks}

\usepackage{multirow}
\usepackage{amsmath,amssymb}
\usepackage{makecell}
\usepackage{array}
\usepackage{wrapfig}
\usepackage{threeparttable}

\providecommand{\cmark}{\ensuremath{\checkmark}}
\providecommand{\xmark}{\ensuremath{\times}}
\definecolor{lightgray}{gray}{0.92}
\newcolumntype{L}[1]{>{\raggedright\arraybackslash}p{#1}}
\newcolumntype{C}[1]{>{\centering\arraybackslash}p{#1}}
\newcolumntype{E}{>{\centering\arraybackslash}m{0.6cm}}
\newcommand{\fullrowgray}{%
  \makebox[0pt][l]{%
    \hspace*{0pt}%
    \raisebox{-0.7ex}[0pt][0pt]{%
      \color{black!8}\rule{\textwidth}{2.8ex}%
    }%
  }%
}
\newcommand{\rowgrayexceptfirsta}[1][0.803\textwidth]{%
  \makebox[0pt][l]{%
    \hspace*{0pt}%
    \raisebox{-0.7ex}[0pt][0pt]{%
      \color{black!8}\rule{#1}{2.8ex}%
    }%
  }%
}
\newcommand{\rowgrayexceptfirstb}[1][0.781\textwidth]{%
  \makebox[0pt][l]{%
    \hspace*{0pt}%
    \raisebox{-0.7ex}[0pt][0pt]{%
      \color{black!8}\rule{#1}{2.8ex}%
    }%
  }%
}

\newcommand{\algo}{STAR\xspace}

\title{STAR: Steering Action Representations with Proprioception for VLA Inference}

\author{%
  \normalfont
  \textbf{Guangyu Zhao}\textsuperscript{1,*} \quad
  \textbf{Yongchang Zhang}\textsuperscript{1,*} \quad
  \textbf{Haoxuan Ru}\textsuperscript{1,*} \quad
  \textbf{Kaichen He}\textsuperscript{1} \\
  \textbf{Haoran Wen}\textsuperscript{2} \quad
  \textbf{Kun Zhan}\textsuperscript{2} \quad
  \textbf{Zeyu Zhang}\textsuperscript{3} \quad
  \textbf{Yin Li}\textsuperscript{4} \quad
  \textbf{Yiwu Zhong}\textsuperscript{1,\textdagger} \\[3pt]
  \textsuperscript{1}Peking University \quad
  \textsuperscript{2}Li Auto \quad
  \textsuperscript{3}BIGAI \quad
  \textsuperscript{4}University of Wisconsin--Madison \\[3pt]
  \textsuperscript{*}Equal contribution. \quad
  \textsuperscript{\textdagger}Corresponding author.
}

\begin{document}

\maketitle

\begin{abstract}

Vision-language-action models (VLAs) map language instructions and visual observations to robot actions, showing strong capabilities across robotic manipulation tasks.
However, in our probing experiment, we find that action representations near the action output do not fully preserve task-relevant visual grounding.
Motivated by this late-layer grounding gap, we introduce \algo, for \textbf{ST}eering \textbf{A}ction \textbf{R}epresentations, a training-free inference method for pre-trained VLAs.
\algo applies an exponential moving average over late-layer residual updates, allowing later action representations to inherit object-aware update directions from earlier layers without model updates or architectural changes.
It further uses recent proprioceptive states to adapt the mixing strength, using conservative mixing when execution dynamics change rapidly.
Experiments on robot manipulation benchmarks show that \algo improves late-layer object awareness and increases average task success rates by 4.0 percentage points.
Our code is available at \url{https://github.com/PKU-VaLuE-Lab/STAR}.
\end{abstract}

\section{Introduction}
\label{sec:introduction}

Despite the tremendous success of vision-language models (VLMs)~\citep{llava,molmo,prismatic,eaglevlm,paligemma,qwen3},
previous work has shown that they may overly rely on language priors, producing responses that are not fully grounded in the visual input~\cite{pope,vcd,opera,m3id,damo}.
Meanwhile, recent studies also observe that vision-language-action models (VLAs) exhibit similar behavior patterns, such as insufficient use of task-relevant visual cues~\cite{reconvla,gaze_regularized_vla,obeyed_vla,pointvla,coavla} or distraction from irrelevant visual tokens~\cite{dtp,cgvd,tag,cogvla}.
Inspired by these works, we ask a key question: {\textit{how strongly are action representations in VLAs grounded in task-relevant visual evidence}}?

\begin{figure}[t]
    \centering
    \includegraphics[width=0.8\linewidth]{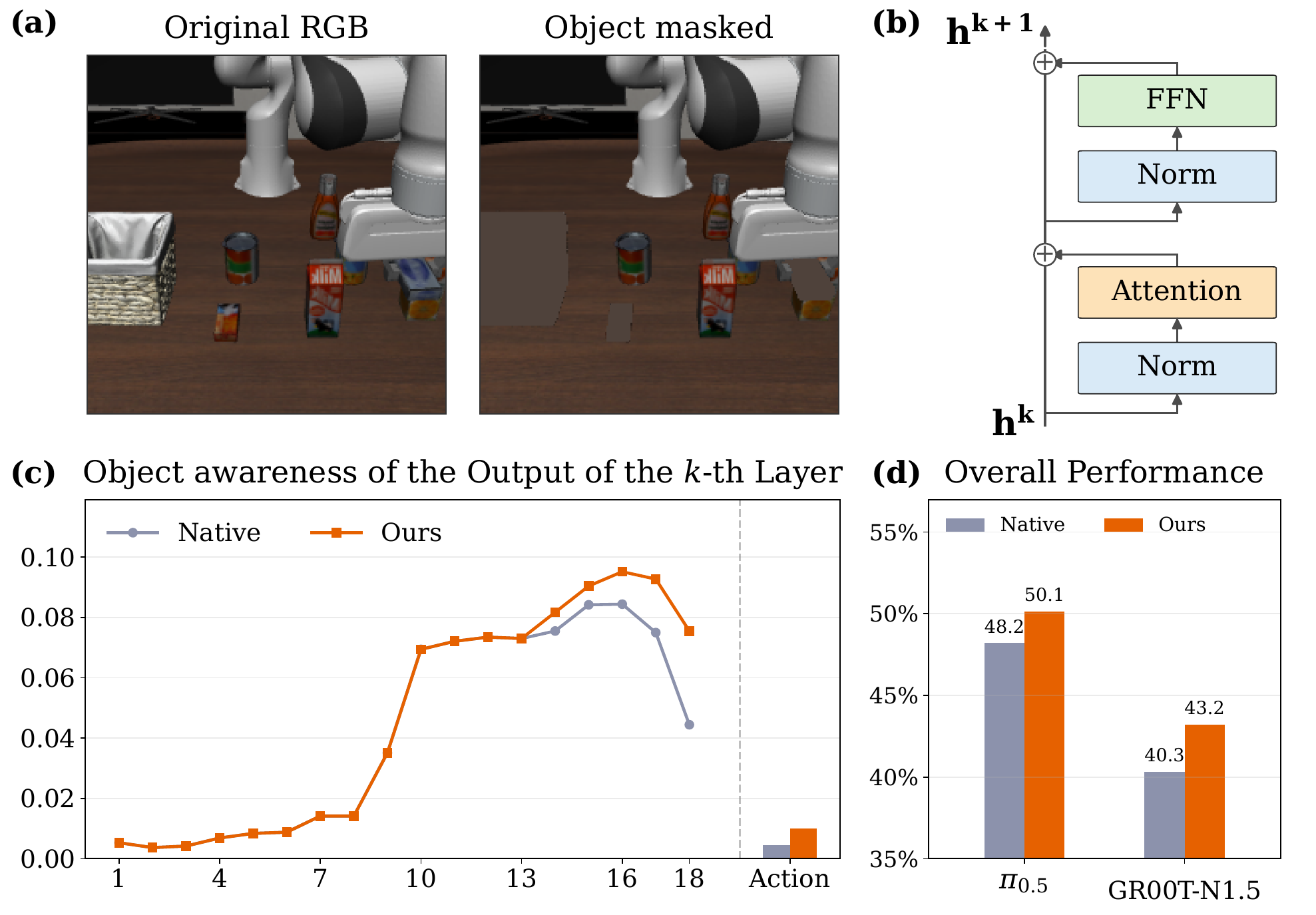}
    \caption{
    \textbf{A case study of layer-wise object awareness in VLA inference.}
    \textbf{(a)} Original and object-masked RGB observations in LIBERO-PRO, where task-relevant object regions are replaced with the mean RGB values of the image.
    \textbf{(b)} A Transformer layer in a VLA, mapping the action hidden state from $h^k$ to $h^{k+1}$ through a residual update.
    \textbf{(c)} Object awareness of $\pi_{0.5}$ layer outputs across depth under native inference and our method \algo.
    \textbf{(d)} Success rates of native inference and our method \algo on LIBERO-PRO for $\pi_{0.5}$ and GR00T-N1.5, respectively.
    }
    \label{fig:motivation}
\end{figure}

To probe the dependence of action representations on visual evidence, we design a controlled experiment.
As illustrated in Figure~\ref{fig:motivation} (a), we mask the target object pixels by replacing them with the mean RGB values of the image.
By masking the visual information important for action execution, we expect the action representations to differ from those produced without interference.
To measure the representation change induced by object masking, we define an \emph{object awareness} score, computed as $\Vert h^k_{\mathrm{m}} - h^k_{\mathrm{o}}\Vert\ /\ \Vert h^k_{\mathrm{o}}\Vert$. Here, $h^k$ is the hidden state of the $k$-th layer in Figure~\ref{fig:motivation}(b), while $\mathrm{o}$ and $\mathrm{m}$ denote the original and masked observations, respectively.
This metric provides a heuristic estimate of how strongly task-relevant visual evidence affects the action representation.
A larger score indicates a larger difference in the representation after masking.
The resulting object-awareness score for $\pi_{0.5}$~\citep{pi05} is reported in Figure~\ref{fig:motivation} (c).
We find that, during native inference, object awareness increases through early-to-middle layers and drops in late layers.
This suggests that visual evidence is gradually aggregated into action representations from early to middle layers, but fades away in late layers.
This probing experiment demonstrates a grounding gap in late layers during VLA inference.

To address the gap, we introduce \textbf{\algo}, for \textbf{ST}eering \textbf{A}ction \textbf{R}epresentations, a training-free method that regularizes the action representations in late layers without updating model weights.
Intuitively, \algo steers late-layer action representations by carrying residual updates from earlier layers to later layers.
Specifically, \algo applies an exponential moving average (EMA)~\citep{ema} over late-layer updates.
When action semantics change rapidly, such as during end-effector rotation or fine-grained adjustment, the intervention should be conservative.
Conversely, when the robot does not rotate, such as when moving in a straight line, the intervention can be stronger.
These differences in execution dynamics are reflected in recent proprioceptive trajectories.
In short, \algo uses a proprioception-modulated EMA to smooth cross-layer residual updates, carrying object-aware directions from earlier layers.

This design is validated through our \emph{object awareness} measurement and robotic manipulation benchmarks. In the test of object awareness, as shown in Figure~\ref{fig:motivation} (c), \algo increases the object awareness of late-layer action representations.
In robotic manipulation benchmarks, \algo enhances the performance of $\pi_{0.5}$~\citep{pi05} and GR00T-N1.5~\citep{gr00t-n15} in LIBERO-PRO, as shown in Figure~\ref{fig:motivation} (d), and yields an average success-rate improvement of 4.0 percentage points in our main experiments.

Our contributions are summarized as follows:

\textbullet\ \ We introduce \algo, a training-free VLA inference method that regularizes action representations toward better visual grounding.

\textbullet\ \ Our technical innovation is an EMA that combines residual updates across late layers, together with a modulation strategy that adjusts its coefficient using proprioceptive states.

\textbullet\ \ Extensive experiments validate that \algo can consistently improve pre-trained VLA models across benchmarks, with an average success-rate improvement of 4.0 percentage points.

\section{Related Work}
\label{sec:related}

\subsection{Vision-Language-Action Models}
Vision-Language-Action models (VLAs) are robot policies adapted from pretrained vision-language models (VLMs) with strong multimodal understanding abilities~\citep{rt2,vlm4vla,vlasurvey,vlasurvey_realworld_app,vlasurvey_concept}.
RT-2~\cite{rt2} established this paradigm by co-fine-tuning a VLM on vision-language data and robot trajectories, thereby aligning it with robotic control. This line was further developed by RoboFlamingo~\cite{roboflamingo} and OpenVLA~\cite{openvla}, which showed that VLM backbones can serve as strong and scalable policy priors. More recent works extend the action-token formulation~\citep{vlasurvey_token} or the training pipeline while retaining this VLM-centric recipe.
$\pi_0$~\cite{pi0} and $\pi_{0.5}$~\cite{pi05} adopt flow-based continuous action modeling.
UniVLA~\cite{univla} studies cross-embodiment policy learning through task-centric latent actions. The GR00T series~\cite{gr00t-n1,gr00t-n15,gr00t-n16,gr00t-n17} balances efficiency and performance with a dual-system design.
ReconVLA~\cite{reconvla} strengthens the perceptual role of VLA models by using reconstructive supervision.
As VLA foundation models improve, our work explores a complementary training-free direction that further enhances their inference-time performance without additional data collection or parameter updates.

\subsection{Training-free Methods in Embodied Policies}
A growing body of work improves embodied policies at test time without updating model weights, with existing approaches differing mainly in where the intervention is applied.
One line of work modifies \textbf{diffusion or flow trajectories}. TNB~\cite{tothenoiseandback}, HAJL~\cite{hajl}, and DemoDiffusion~\cite{demodiffusion} re-noise and denoise alternative or retargeted action sequences, effectively blending them with the generative prior of a pre-trained policy. VLS~\cite{vls} and DynaGuide~\cite{dynaguide} further steer denoising with guidance from additional vision foundation models.
Another line targets \textbf{action decoding}. BID~\cite{bid} restores reactivity via closed-loop resampling over candidate chunks. RTC~\cite{rtc} enables asynchronous real-time control by freezing committed actions and inpainting the remaining chunk. Recent test-time scaling methods use online search or candidate selection to improve action choice, including VLA-Reasoner~\cite{vlareasoner} and anti-exploration steering~\cite{taco}.
A more recent line intervenes at the \textbf{spatial visual-token} level. IVRA~\cite{ivra} refines relations among visual tokens using encoder-derived affinity hints, CGVD~\cite{cgvd} suppresses spatial distractors through concept-gated visual distillation, and DTP~\cite{dtp} prunes task-irrelevant visual tokens that mislead action generation.
Overall, these results highlight that performance gains can be achieved purely through inference-time modifications.
Most approaches operate on explicit input-output interfaces.
In contrast, our work explores whether directly intervening on transformer hidden-state dynamics can improve visual grounding and downstream control.

\subsection{Visual Hallucination}
Visual hallucination is common in multimodal models, including VLMs~\citep{pope,haelm,hallusionbench,amber,mllm_hallucination_survey} and embodied visuomotor policies, where it manifests as weak visual grounding or action-level hallucination~\citep{reconvla,pointvla,action_hallucination}.
For VLMs, DAMO~\cite{damo} mitigates visual hallucination by accumulating activation momentum during decoding. Similarly, DCLA~\cite{tang2025dcla} and VASparse~\cite{zhuang2025vasparse} suggest that inference-time control of hidden states or visual token flow can improve faithfulness without retraining.
For embodied policies, the analogous failure is inaccurate visual grounding, such as attending to distractors or acting on the wrong object. Knowledge Insulation~\cite{ki} shows that naively adapting VLM backbones to action can erode the transferred semantic knowledge that supports grounding. ReconVLA~\cite{reconvla} strengthens grounding through a reconstructive objective over gaze regions, whereas DTP~\cite{dtp} offers a lightweight training-free alternative by pruning distracting tokens at inference time.
Our method is closest in spirit to DAMO, as both methods use training-free control of layer-wise dynamics. However, DAMO targets VQA tasks in VLMs, while our work targets robotic action execution. This embodied setting introduces proprioceptive execution states as a key signal for controlling the intervention, a feature specific to robot policies.

\section{STAR: Steering Action Representations during Inference}
\label{sec:method}

\subsection{Overview}

Vision-language-action models generate actions from visual observations, proprioceptive states, and language instructions.
At each control step $t$, a VLA policy receives an observation
$o_t=(I_t,s_t)$, where $I_t$ denotes the visual input and $s_t$ denotes the robot proprioceptive state, together with a language instruction $x$. The policy outputs an action chunk: 
\begin{equation}
    A_t = [a_{t, 1}, a_{t, 2}, \ldots, a_{t, H_p}],
\end{equation}
where $H_p$ is the prediction horizon, and the first $H_e$ actions $a_{t, 1:H_e}$ are executed.
Each executed action is decoded from the model's final action representations, making it important to preserve task-relevant visual evidence until the late layers.

\begin{wrapfigure}{r}{0.48\textwidth}
    \vspace{-8pt}
    \centering
    \includegraphics[width=0.46\textwidth]{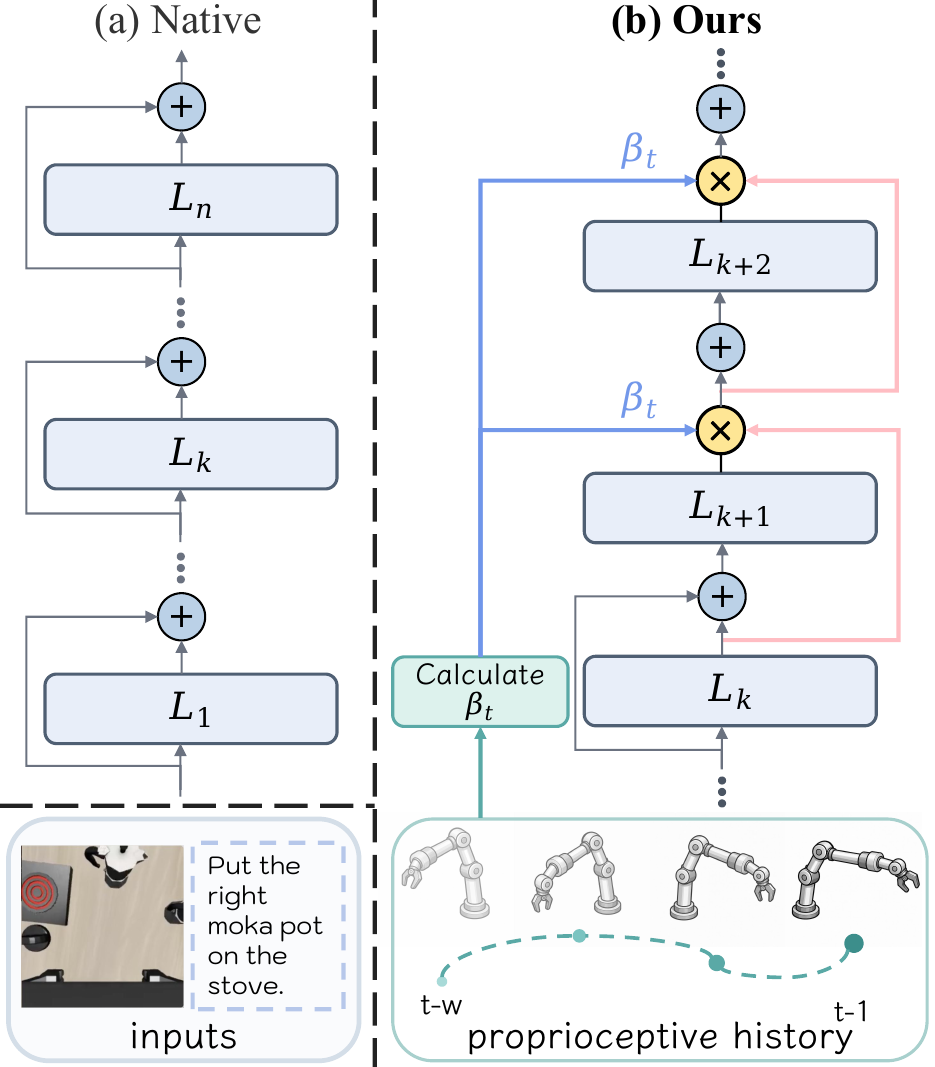}
    \caption{
    \textbf{Overview of action generation at step $t$.}
    \algo estimates a step-level EMA coefficient $\beta_t$ from proprioceptive history and applies it to layer-wise updates.
    Blue arrows denote the EMA coefficient $\beta_t$ derived from proprioceptive history, while pink arrows denote the cross-layer update momentum used in the EMA mixing.
    }
    \label{fig:method_overview}
    \vspace{5pt}
\end{wrapfigure}

Our key design principle is to preserve object-aware update directions from earlier layers and carry them to later layers using EMA.
A fixed EMA strength, however, ignores the non-uniform dynamics of robotic manipulation.
We characterize execution dynamics along a spectrum from \emph{progress}, where the robot moves steadily along a consistent direction, to \emph{transition}, where execution dynamics can change rapidly due to contact changes, end-effector reorientation, or fine-grained adjustment.
Our heuristic is that transition phases require a more conservative EMA coefficient to limit the extent of inference-time editing, thereby preventing the policy from deviating too much from its original behavior.
Following this principle, we propose \algo, a training-free VLA inference method that carries residual updates from earlier, object-aware layers to late layers via a proprioception-modulated EMA. An overview of our method is illustrated in Figure~\ref{fig:method_overview}.

\subsection{EMA-Based Layer-wise Residual Update}

We assume the VLA backbone contains $n$ Transformer layers $L_1,\ldots,L_n$. We define the late-layer mixing region by a relative starting depth $\rho$:
\begin{equation}
    p = \lfloor \rho n \rfloor,
    \qquad
    \mathcal{L}_{\mathrm{early}} = \{L_1,\ldots,L_p\},
    \qquad
    \mathcal{L}_{\mathrm{mix}} = \{L_{p+1},\ldots,L_n\}.
\end{equation}
In our experiments, we set $\rho=2/3$, so mixed updates are applied to the final third of the Transformer layers.
This choice is motivated by the diagnostic study in Figure~\ref{fig:motivation} and prior work on inference-time modification of multimodal models~\citep{damo,aim}. The first $p = \lfloor \rho n \rfloor$ layers are left unchanged, allowing the backbone to form language-conditioned visual evidence. For layers in $\mathcal{L}_{\mathrm{mix}}$, which are closer to action decoding and may become less object-aware, we apply proprioceptively modulated EMA updates.

Formally, we denote the direct output of the $k$-th Transformer block at robot time step $t$ as $\Delta h_t^k$. This is the residual update of hidden state $h_t^k$: 
\begin{equation}
    h_t^k = h_t^{k-1} + \Delta h_t^k.
\end{equation}
STAR mixes residual updates across late layers using an EMA. Analogous to momentum in optimization, it maintains a residual-update trend (``momentum'') $m_t^k$ across the late layers of each forward pass. At control step $t$, this trend is updated using coefficient $\beta_t$:
\begin{equation}
\label{eq:ema_update}
    m_t^k =
    \beta_t m_t^{k-1}
    +
    (1-\beta_t)\Delta h_t^k,
    \qquad
    L_k\in\mathcal{L}_{\mathrm{mix}}.
\end{equation}
% The mixed hidden state is then used as the output of the late layer:
The hidden states of layers in $\mathcal L_{\text{mix}}$ are updated using $m_t$ rather than the original $\Delta h_t$: 
\vspace{1.5\baselineskip}
\begin{equation}
    \hat h_t^k =
    h_t^{k-1} + m_t^k,
    \qquad
    h_t^k \leftarrow \hat h_t^k,
    \quad
    L_k\in\mathcal{L}_{\mathrm{mix}}.
\end{equation}
This operation carries a smoothed update direction across depth, allowing the object-aware information formed in earlier layers to persist through later layers.

Another question is how to choose $\beta_t$ at each control step $t$. A fixed coefficient $\beta_t$ cannot account for how robot execution alternates between \emph{progress} and \emph{transition} phases. We therefore compute $\beta_t$ from causal proprioceptive evidence.

\subsection{Causal Proprioceptive Evidence}

A VLA policy operates in a closed sensorimotor loop, so the robot's own motion provides direct evidence of the current execution dynamics. Specifically, let
$p_t=\mathrm{pos}(s_t)\in\mathbb{R}^3$ denote the end-effector position extracted from $s_t$, and let $R_t=\mathrm{rot}(s_t)$ denote the corresponding orientation representation. We construct four causal quantities from the end-effector trajectory and orientation~history.

First, we measure directional continuity:
\begin{equation}
    c_t =
    \cos(p_t-p_{t-1},\, p_{t-1}-p_{t-2}).
\end{equation}
A high $c_t$ indicates that the end-effector continues to move along a consistent direction, which often corresponds to steady progress.

Second, we measure the motion scale:
\begin{equation}
    v_t = \|p_t-p_{t-1}\|_2.
\end{equation}
This quantity reflects whether the robot is still undergoing substantial translational motion.

Third, we measure an acceleration-like change in translation:
\begin{equation}
    j_t =
    \|(p_t-p_{t-1})-(p_{t-1}-p_{t-2})\|_2.
\end{equation}
This quantity captures abrupt changes in motion, which often occur near transitions, contact changes, or local corrections.

Fourth, we measure orientation change:
\begin{equation}
    \omega_t = d_{\mathrm{rot}}(R_t,R_{t-1}),
\end{equation}
where $d_{\mathrm{rot}}(\cdot,\cdot)$ denotes a rotation distance. Orientation changes are particularly informative for execution dynamics involving grasping, insertion, drawer interaction, or precise placement.

All quantities are computed only from observations available up to the current control step. Therefore, the modulation signal is strictly causal and does not use future states or oracle task information.

\subsection{Causal Normalization}

The raw magnitudes of proprioceptive quantities vary across embodiments, tasks, and simulators. We therefore normalize each scalar using a causal sliding window. For $x_t\in\{c_t,v_t,j_t,\omega_t\}$, we define the window length and the corresponding history as
\begin{equation}
    W = f_{\mathrm{ctrl}}T_w,
    \qquad
    \mathcal{H}_t^x = \{x_{t-W},\ldots,x_{t-1}\},
\end{equation}
where $f_{\mathrm{ctrl}}$ is the control frequency and $T_w$ is the temporal span of the statistics. The normalized value is then computed as
\begin{equation}
    z_t^x =
    \frac{x_t-\mu(\mathcal{H}_t^x)}
    {\sigma(\mathcal{H}_t^x)+\epsilon}.
\end{equation}
Since $\mathcal{H}_t^x$ only contains values before the current control step, the normalization remains strictly causal. This makes the normalized proprioceptive evidence depend on recent execution context rather than on the entire episode history.

\subsection{Proprioceptive Mixing Score}

We compute a signed proprioceptive mixing score from the normalized proprioceptive indicators:
\begin{equation}
    r_t =
    w_c z_t^c
    + w_v z_t^v
    + w_j z_t^j
    + w_\omega z_t^\omega .
\end{equation}
Positive $r_t$ indicates that stronger mixed updates are preferred, while negative $r_t$ indicates that late layers should adapt more strongly to the current update. The sign of each coefficient determines whether the corresponding signal increases or decreases the late-layer cross-layer mixing strength. In our implementation, directional continuity and motion scale provide positive evidence for steady progress and thus stronger mixed updates, whereas abrupt translational change and orientation change provide negative evidence when they indicate transitions or fine adjustments.

To make the scale of $r_t$ stable across tasks, we further apply a robust causal normalization. Let $\mathcal{R}_t=\{r_{t-W},\ldots,r_{t-1}\}$ be the recent history of proprioceptive mixing scores. We compute:
\begin{equation}
    \hat r_t =
    \frac{
        r_t-\operatorname{median}(\mathcal{R}_t)
    }{
        \operatorname{IQR}(\mathcal{R}_t)+\epsilon
    }.
\end{equation}
The use of the median and interquartile range reduces sensitivity to transient spikes in proprioceptive~measurements.

\subsection{Proprioception-Conditioned Mixing Coefficient}

The normalized proprioceptive score is mapped to the continuous coefficient used in the EMA mixed~updates:
\begin{equation}
    \beta_t =
    \beta_{\min}
    +
    (\beta_{\max}-\beta_{\min})
    \sigma(\kappa \hat r_t),
\end{equation}
where $\sigma(\cdot)$ is the sigmoid function and $\kappa$ controls the sharpness of the transition. A larger $\beta_t$ preserves more accumulated update momentum in Eq.~(\ref{eq:ema_update}), helping late layers maintain object-aware update information during steady execution. A smaller $\beta_t$ makes the cross-layer mixed update rely more on the current layer update, which is useful during transitions, contact changes, or fine adjustment.

This operation does not modify the attention map. Instead, it only changes how late-layer residual updates~\citep{resnet,transformer,gpt2} are mixed before being propagated to the action head. Therefore, \algo directly targets the update stream where late-layer visual degradation occurs, while the proprioceptive coefficient $\beta_t$ adapts the cross-layer mixing strength to the robot's current execution dynamics.

\section{Experiments and Results}
\label{sec:experiments}

We evaluate \algo on two manipulation benchmarks, SimplerEnv~\citep{simplerenv} and LIBERO-PRO~\citep{liberopro}. These benchmarks cover complementary aspects of VLA control: SimplerEnv emphasizes Google Robot-style tabletop manipulation with diverse object and articulated-object interactions, while LIBERO-PRO focuses on long-horizon desktop manipulation with compositional task variations. Our experiments are designed to answer three questions:
(i) whether \algo improves frozen VLA policies without additional training,
(ii) whether the proposed cross-layer residual mixing mechanism transfers across different VLA backbones, and
(iii) which design choices are responsible for the improvement, including the proprioceptive evidence, beta scheduling, and intervention start layer.

\subsection{Implementation Details}
\label{subsec:implementation_details}

\textbf{Backbones.}
We evaluate \algo on three representative VLA backbones with different vision-language architectures and action-generation heads.
\textbf{GR00T-N1.5}~\citep{gr00t-n15} uses an NVIDIA Eagle VLM to encode language instructions and visual observations, and conditions a flow-based action head on the resulting vision-language embeddings together with robot state and action tokens.
\textbf{OpenVLA}~\citep{openvla} is an open-source VLA model built on a Prismatic-style VLM. It combines DINOv2 and SigLIP visual encoders with a projection module and a Llama-2-based language backbone. We use OpenVLA-7B in our experiments.
\textbf{$\mathbf{\pi_{0.5}}$}~\citep{pi05} follows a PaliGemma-style VLM plus action-expert design. Its VLM backbone uses a SigLIP vision encoder and a Gemma-based language transformer, and the action expert generates continuous action chunks through flow matching.
For each backbone, we compare the original frozen policy with the same policy equipped with \algo. All model weights remain frozen during evaluation.

\textbf{Evaluation protocol.}
We apply \algo only at inference time and follow the official evaluation protocol of each benchmark. We report task success rate as the primary metric, including both per-task success rates and average success rates.

\subsection{Benchmarks}
\label{subsec:benchmarks_metrics}

\textbf{SimplerEnv.}
SimplerEnv~\citep{simplerenv} evaluates policies in simulated Google Robot manipulation environments. We use six Fractal/Google Robot tasks.
\textit{Open Drawer} requires the robot to grasp and pull an articulated drawer handle.
\textit{Place in Drawer} requires opening a drawer, placing an object inside, and completing the placement under occlusion and articulated-object constraints.
\textit{Pick Coke Can} evaluates object-centric grasping and lifting for a commonly used can object.
\textit{Move Near} tests spatial relation reasoning by asking the robot to move one object close to another target object.
\textit{Close Drawer} requires pushing or sliding the drawer back to a closed state after interaction.
\textit{Pick Object} evaluates generic object picking across randomized object instances and placements. These tasks jointly cover grasping, articulated-object interaction, spatial relocation, and relation-aware control.

\textbf{LIBERO-PRO.}
LIBERO-PRO~\citep{liberopro} evaluates long-horizon manipulation under multiple forms of generalization. We report five evaluation splits.
\textit{LIBERO-10-lan} paraphrases instructions while preserving task semantics.
\textit{LIBERO-10-object} modifies non-essential object attributes such as color, texture, or size.
\textit{LIBERO-10-swap} perturbs initial object placements and spatial configurations.
\textit{LIBERO-10-task} changes target objects, action requirements, or task goals.
\textit{LIBERO-10-env} varies visual context such as backgrounds, lighting, or textures. Each split contains ten tasks, and we report the success rate averaged over the tasks within each split. For STAR with $\pi_{0.5}$, we evaluate 50 episodes per task, totaling 500 episodes per split and 2,500 episodes across the five splits.

\subsection{Main Results}
\label{subsec:main_results}

We compare \algo with native inference and a training-free attention-side baseline, DTP~\citep{dtp}. DTP aims to improve VLA inference by identifying task-relevant visual regions and pruning the distraction image tokens according to action-token attention. It is therefore a natural baseline for testing whether improving visual attention allocation is sufficient for more reliable action generation. However, DTP requires an additional forward pass to obtain the attention signals used for pruning, whereas \algo performs single-pass inference and directly modulates the late residual update stream without changing tokens, attention maps, or model weights.

\begin{table*}[t]
\centering
\setlength{\tabcolsep}{3.5pt}
\renewcommand{\arraystretch}{1.18}
\begin{threeparttable}
\caption{
\textbf{Main results on SimplerEnv.}
For OpenVLA, the average is computed over the available tasks. \algo improves frozen VLA policies on both backbones and achieves the best average success rate, outperforming both native inference and DTP while remaining single-pass.
}
\label{tab:simplerenv_main}

\begin{tabular}{@{}l l c c c c c c c@{}}
\toprule
\multirow{2}{*}{\textbf{Backbone}} &
\multirow{2}{*}{\textbf{Method}} &
\multicolumn{6}{c}{\textbf{SimplerEnv Google Robot Tasks}} &
\multirow{2}{*}{\textbf{Avg.}} \\
\cmidrule(lr){3-8}
& &
\makecell[c]{Open\\Drawer} &
\makecell[c]{Place in\\Drawer} &
\makecell[c]{Pick\\Coke Can} &
\makecell[c]{Move\\Near} &
\makecell[c]{Close\\Drawer} &
\makecell[c]{Pick\\Object} &
\\
\midrule

\multirow{3}{*}{GR00T-N1.5~\citep{gr00t-n15}}
& Native
& 26.7\% & 4.0\% & 69.3\% & 68.7\% & 45.0\% & 32.3\% & 41.0\% \\
& DTP
& 26.7\% & 4.0\% & 59.3\% & 68.7\% & 41.7\% & 39.3\% & 40.0\% \\
& \rowgrayexceptfirsta\textbf{\algo (Ours)}
& \textbf{32.0\%} & \textbf{4.7\%} & \textbf{77.3\%} & \textbf{76.7\%} & \textbf{47.3\%} & \textbf{40.0\%} & \textbf{46.3\%} \\
\midrule

\multirow{3}{*}{OpenVLA~\citep{openvla}}
& Native
& 7.4\% & 0.0\% & 65.3\% & 56.7\% & 22.2\% & -- & 30.3\% \\
& DTP
& 11.1\% & 0.0\% & 69.3\% & \textbf{69.0\%} & 22.2\% & -- & 34.3\% \\
& \rowgrayexceptfirsta\textbf{\algo (Ours)}
& \textbf{18.5\%} & \textbf{3.7\%} & \textbf{70.0\%} & 61.7\% & \textbf{25.9\%} & -- & \textbf{36.0\%} \\
\bottomrule
\end{tabular}
\end{threeparttable}
\vspace{-2mm}
\end{table*}

\begin{table*}[!t]
\centering
\setlength{\tabcolsep}{7.7pt}
\renewcommand{\arraystretch}{1.18}
\caption{
\textbf{Main results on LIBERO-PRO.}
\algo improves the average success rate on both GR00T-N1.5 and $\pi_{0.5}$, while DTP gives limited or mixed gains on this long-horizon benchmark.
}
\label{tab:libero_main}

\begin{tabular*}{\textwidth}{@{\extracolsep{\fill}} l l c c c c c c @{}}
\toprule
\multirow{2}{*}{\textbf{Backbone}} &
\multirow{2}{*}{\textbf{Method}} &
\multicolumn{5}{c}{\textbf{LIBERO-PRO Perturbations on LIBERO-10}} &
\multirow{2}{*}{\textbf{Avg.}} \\
\cmidrule(lr){3-7}
& &
\makecell[c]{lan} &
\makecell[c]{object} &
\makecell[c]{swap} &
\makecell[c]{task} &
\makecell[c]{env} &
\\
\midrule

\multirow{3}{*}{GR00T-N1.5~\citep{gr00t-n15}}
& Native
& 81.2\% & 40.4\% & 0.0\% & 6.0\% & 73.8\% & 40.3\% \\
& DTP
& 82.0\% & 42.2\% & 0.0\% & 1.8\% & 75.6\% & 40.3\% \\
& \rowgrayexceptfirstb\textbf{\algo (Ours)}
& \textbf{82.8\%} & \textbf{48.0\%} & \textbf{2.2\%} & \textbf{6.8\%} & \textbf{76.4\%} & \textbf{43.2\%} \\
\midrule

\multirow{3}{*}{$\pi_{0.5}$~\citep{pi05}}
& Native
& \textbf{92.8\%} & 66.2\% & \textbf{9.2\%} & 6.8\% & 65.8\% & 48.2\% \\
& DTP
& 90.4\% & 64.6\% & 8.0\% & 7.8\% & \textbf{67.6\%} & 47.7\% \\
& \rowgrayexceptfirstb\textbf{\algo (Ours)}
& 92.6\% & \textbf{67.0\%} & 7.2\% & \textbf{18.8\%} & 65.0\% & \textbf{50.1\%} \\
\bottomrule
\end{tabular*}
\vspace{-2mm}
\end{table*}

Table 1 summarizes the SimplerEnv results. STAR improves the average success rate from 41.0\% to 46.3\% on GR00T-N1.5 and from 30.3\% to 36.0\% on OpenVLA. The gains are consistent across all six GR00T-N1.5 tasks and are stronger overall than those of DTP, showing that STAR is effective across the two backbone-specific evaluation settings. Table 2 reports the LIBERO-PRO results. STAR improves the average success rate from 40.3\% to 43.2\% on GR00T-N1.5 and from 48.2\% to 50.1\% on $\pi_{0.5}$. GR00T-N1.5 gains most on the object split. For $\pi_{0.5}$, the largest gain is on the task split, from 6.8\% to 18.8\%, while the object split improves slightly and the language, swap, and environment splits decline. These results support the applicability of STAR across both autoregressive-style and continuous-action VLA backbones.

\subsection{Ablation Studies}
\label{subsec:ablation}

We conduct ablation studies with OpenVLA on SimplerEnv, and all ablations follow the same protocol as the main SimplerEnv evaluation. We focus on three factors: the proprioceptive evidence used to compute the proprioceptive mixing score, the beta scheduling strategy, and the layer region where modulation is applied. These ablations isolate the main design choices of \algo and test whether the observed gains come from meaningful proprioceptive modulation rather than from a generic late-layer perturbation.

\subsubsection{Effect of Proprioceptive Evidence}
\label{subsec:ablation_proprio}

\begin{table*}[t]
\centering
\small
\setlength{\tabcolsep}{2.5pt}
\renewcommand{\arraystretch}{1.16}
\caption{
\textbf{Ablation on proprioceptive evidence.}
Using all four proprioceptive cues gives the best average success rate, showing that mixed updates benefit from both steady-progress and transition-sensitive signals.
}
\label{tab:ablation_proprio}

\begin{tabular*}{\textwidth}{
@{\extracolsep{\fill}}
>{\raggedright\arraybackslash}p{3.5cm}
@{\hspace{0.45em}\extracolsep{2.5pt}}
E E E E
@{\hspace{0.75em}\extracolsep{\fill}}
c c c c c c
@{}}
\toprule
\multirow{2}{*}{\textbf{Variant}} &
\multicolumn{4}{c}{\textbf{Proprioceptive Evidence}} &
\multicolumn{5}{c}{\textbf{SimplerEnv Google Robot Tasks}} &
\multirow{2}{*}{\textbf{Avg.}} \\
\cmidrule(lr){2-5}
\cmidrule(lr){6-10}
&
$c_t$ & $v_t$ & $j_t$ & $\omega_t$
& \makecell[c]{Open\\Drawer}
& \makecell[c]{Place in\\ Drawer}
& \makecell[c]{Pick\\Coke Can}
& \makecell[c]{Move\\Near}
& \makecell[c]{Close\\Drawer}
& \\
\midrule

\fullrowgray
\textbf{Full \algo (Ours)}
& \cmark & \cmark & \cmark & \cmark
& \textbf{18.5\%} & \textbf{3.7\%} & 70.0\% & \textbf{61.7\%} & \textbf{25.9\%} & \textbf{36.0\%} \\

\addlinespace[1.0pt]
w/o direction consistency
& \xmark & \cmark & \cmark & \cmark
& 14.8\% & 0.0\% & 64.0\% & 56.7\% & 14.8\% & 30.1\% \\

w/o translation scale
& \cmark & \xmark & \cmark & \cmark
& 11.1\% & 0.0\% & 66.7\% & 55.0\% & 22.2\% & 31.0\% \\

w/o jerk
& \cmark & \cmark & \xmark & \cmark
& 14.8\% & \textbf{3.7\%} & 65.3\% & 58.3\% & 14.8\% & 31.4\% \\

w/o rotation change
& \cmark & \cmark & \cmark & \xmark
& 14.8\% & 0.0\% & 66.7\% & 58.3\% & 18.5\% & 31.7\% \\

\midrule

Positive evidence only
& \cmark & \cmark & \xmark & \xmark
& 3.7\% & 0.0\% & 62.7\% & \textbf{61.7\%} & 18.5\% & 29.3\% \\

Transition evidence only
& \xmark & \xmark & \cmark & \cmark
& 11.1\% & \textbf{3.7\%} & \textbf{70.7\%} & 58.3\% & 22.2\% & 33.2\% \\

No proprioceptive evidence
& \xmark & \xmark & \xmark & \xmark
& 11.1\% & 0.0\% & 56.0\% & 53.3\% & 18.5\% & 27.8\% \\

\bottomrule
\end{tabular*}
\vspace{-2mm}
\end{table*}

Table~\ref{tab:ablation_proprio} studies the role of proprioceptive evidence. Removing any single cue causes a 4.3--5.9 percentage point drop in average success rate, indicating that the four signals provide complementary information for characterizing execution dynamics across tasks. Direction consistency and translation scale help identify steady progress, while jerk and rotation change provide transition-sensitive evidence for contact changes and fine manipulation. The grouped ablations further support this interpretation: using only positive evidence reaches 29.3\%, and using only transition evidence reaches 33.2\%, both below the full model. The no-proprioception variant obtains 27.8\%, confirming that the improvement is not simply caused by adding a late-layer mixing operation, but by conditioning the cross-layer mixing strength on meaningful robot-state dynamics.

\subsubsection{Effect of Beta Scheduling}
\label{subsec:ablation_beta}

We report success rates averaged over eight evaluation runs for this ablation.

\begin{table*}[t]
\centering
\setlength{\tabcolsep}{4.2pt}
\renewcommand{\arraystretch}{1.16}
\caption{
\textbf{Ablation on beta scheduling.}
A proprioceptively scheduled $\beta$ with a moderate range outperforms both fixed cross-layer mixing strengths and poorly calibrated dynamic ranges.
}
\label{tab:ablation_beta}

\begin{tabular*}{\textwidth}{
@{\extracolsep{\fill}}
>{\raggedright\arraybackslash}p{5cm}
c c c c c
c
@{}}
\toprule
\multirow{2}{*}{\textbf{Variant}} &
\multicolumn{5}{c}{\textbf{SimplerEnv Google Robot Tasks}} &
\multirow{2}{*}{\textbf{Avg.}} \\
\cmidrule(lr){2-6}
&
\makecell[c]{Open\\Drawer}
& \makecell[c]{Place in\\ Drawer}
& \makecell[c]{Pick\\Coke Can}
& \makecell[c]{Move\\Near}
& \makecell[c]{Close\\Drawer}
& \\
\midrule

Fixed-$\beta = 0.25$
& 11.1\% & 0.0\% & 68.0\% & 56.7\% & 22.2\% & 31.6\% \\

Fixed-$\beta = 0.35$
& 14.8\% & 0.0\% & 65.3\% & 56.7\% & 14.8\% & 30.3\% \\

Fixed-$\beta = 0.45$
& 9.3\% & 1.9\% & 65.7\% & 58.3\% & 18.5\% & 30.7\% \\

Fixed-$\beta = 0.55$
& 14.8\% & 0.0\% & 64.0\% & 60.0\% & 14.8\% & 30.7\% \\

\midrule

Dynamic-$\beta$ $(0.15,0.55)$
& 14.8\% & 0.0\% & 64.0\% & 55.0\% & 22.2\% & 31.2\% \\

\fullrowgray
\textbf{Dynamic-$\boldsymbol{\beta}$ $\mathbf{(0.25,0.55)}$ (Ours)}
& \textbf{18.5\%} & \textbf{3.7\%} & 70.0\% & 61.7\% & \textbf{25.9\%} & \textbf{36.0\%} \\

Dynamic-$\beta$ $(0.35,0.55)$
& 7.4\% & 0.0\% & \textbf{70.7\%} & 51.7\% & 14.8\% & 28.9\% \\

\midrule

Dynamic-$\beta$ $(0.25,0.45)$
& 11.1\% & 0.0\% & 64.0\% & 53.3\% & \textbf{25.9\%} & 30.9\% \\

\fullrowgray
\textbf{Dynamic-$\boldsymbol{\beta}$ $\mathbf{(0.25,0.55)}$ (Ours)}
& \textbf{18.5\%} & \textbf{3.7\%} & 70.0\% & 61.7\% & \textbf{25.9\%} & \textbf{36.0\%} \\

Dynamic-$\beta$ $(0.25,0.65)$
& \textbf{18.5\%} & \textbf{3.7\%} & 66.7\% & \textbf{63.3\%} & 18.5\% & 34.1\% \\

\bottomrule
\end{tabular*}
\vspace{-2mm}
\end{table*}

Table~\ref{tab:ablation_beta} compares fixed and dynamic choices of the cross-layer mixing coefficient. Fixed-$\beta$ variants reach at most 31.6\% average success, showing that a single cross-layer mixing strength cannot accommodate the different execution states in SimplerEnv. Dynamic scheduling improves performance when the range is properly calibrated: decreasing the lower or upper endpoint yields 31.2\% with Dynamic-$\beta(0.15,0.55)$ and 30.9\% with Dynamic-$\beta(0.25,0.45)$, respectively, while increasing them yields 28.9\% with Dynamic-$\beta(0.35,0.55)$ and 34.1\% with Dynamic-$\beta(0.25,0.65)$. The proposed range, Dynamic-$\beta(0.25,0.55)$, achieves the best average success rate of 36.0\%, showing that both adaptivity and range calibration are necessary for effective proprioception-conditioned mixed updates.

\subsubsection{Effect of Start Layer}
\label{subsec:ablation_start_layer}

\begin{table*}[t]
\centering
\setlength{\tabcolsep}{4.0pt}
\renewcommand{\arraystretch}{1.18}
\caption{
\textbf{Ablation on start layer.}
Applying mixed updates only to the final third of layers gives the best average success rate, supporting late residual updates as the effective intervention target.
}
\label{tab:ablation_start_layer}

\begin{tabular*}{\textwidth}{
@{\extracolsep{\fill}}
>{\raggedright\arraybackslash}p{2.8cm}
c c c c c
c
@{}}
\toprule
\multirow{2}{*}{\textbf{Variant}} &
\multicolumn{5}{c}{\textbf{SimplerEnv Google Robot Tasks}} &
\multirow{2}{*}{\textbf{Avg.}} \\
\cmidrule(lr){2-6}
&
\makecell[c]{Open\\Drawer}
& \makecell[c]{Place in\\ Drawer}
& \makecell[c]{Pick\\Coke Can}
& \makecell[c]{Move\\Near}
& \makecell[c]{Close\\Drawer}
& \\
\midrule

$\rho=0$
& 14.8\% & 0.0\% & 49.3\% & 53.3\% & 11.1\% & 25.7\% \\

$\rho=1/3$
& 11.1\% & 0.0\% & 55.0\% & \textbf{68.3\%} & 3.7\% & 27.6\% \\

\fullrowgray
\textbf{$\boldsymbol{\rho}$ $\mathbf{=2/3}$ (Ours)}
& \textbf{18.5\%} & \textbf{3.7\%} & \textbf{70.0\%} & 61.7\% & \textbf{25.9\%} & \textbf{36.0\%} \\

$\rho=1$ (Native)
& 7.4\% & 0.0\% & 65.3\% & 56.7\% & 22.2\% & 30.3\% \\

\bottomrule
\end{tabular*}
\vspace{-2mm}
\end{table*}

Table~\ref{tab:ablation_start_layer} evaluates which layer region should use mixed updates. Mixing from the first layer reduces the average success rate to 25.7\%, suggesting that early visual-language computation should remain unchanged. Starting from the middle of the stack improves \textit{Move Near} to 68.3\%, but weakens object-centric and articulated-object manipulation, leading to a 27.6\% average. Applying mixed updates only to the final third of layers achieves the best average success rate of 36.0\% and the strongest performance on four out of five tasks. This supports the central design of \algo: intervene on the late residual update stream closest to action decoding, while preserving the backbone's native early-to-middle multimodal processing.

\subsection{Real-Robot Experiments}
\label{subsec:real_robot}

We further evaluate \algo with $\pi_{0.5}$ on a real-robot task that requires picking up a pear and placing it in a bowl. We collect 100 episodes and fine-tune $\pi_{0.5}$ for 15,000 steps. Both native inference and \algo use the same fine-tuned checkpoint, with \algo applied only at inference time. Table~\ref{tab:real_robot} reports the success rates over 40 evaluation trials.

\begin{table}[h]
\centering
\setlength{\tabcolsep}{12pt}
\renewcommand{\arraystretch}{1.15}
\caption{
\textbf{Real-robot results.}
Success rates on the pear-to-bowl task with $\pi_{0.5}$.
}
\label{tab:real_robot}
\begin{tabular}{l c c}
\toprule
\textbf{Method} & \textbf{Successes / Trials} & \textbf{Success Rate} \\
\midrule
Native & $17/40$ & 42.5\% \\
\textbf{\algo (Ours)} & $19/40$ & \textbf{47.5\%} \\
\bottomrule
\end{tabular}
\vspace{-2mm}
\end{table}

On the real-robot system with a single NVIDIA RTX 3090, \algo introduces only 3.3\% additional inference latency relative to native inference.

\section{Conclusion}
\label{sec:conclusion}

We introduced \algo, a training-free inference method that improves visual grounding in pre-trained VLA models by steering late-layer action representations. Based on the observation that object awareness weakens near the action output, \algo applies an EMA over late-layer residual updates to preserve object-aware update directions from earlier layers. It further adapts the mixing strength using recent proprioceptive states, making the intervention more conservative when action semantics change rapidly. Experiments show that \algo improves late-layer object awareness and consistently raises task success rates across robotic manipulation benchmarks. Our results highlight training-free residual steering as an effective and complementary way to enhance VLA foundation models without additional training.

\textbf{Limitations.}
While STAR improves training-free VLA inference across multiple benchmarks and backbones, several limitations remain. 
First, STAR requires access to intermediate Transformer hidden states. It therefore cannot be directly applied to black-box policies that expose only action outputs.
Second, the proprioceptive cues used by STAR are designed primarily for single-end-effector manipulation tasks. Extending this design to more complex embodiments remains to be explored.
Finally, STAR is based on an empirical observation that sensitivity to task-relevant objects declines in later VLA Transformer layers. We observe this pattern in the models evaluated here, but it is not theoretically guaranteed for all pretrained VLA models. STAR may offer limited benefits when this pattern is absent.

\bibliographystyle{plainnat}
\bibliography{references}

@inproceedings{rt2,
  title={{RT-2}: Vision-Language-Action Models Transfer Web Knowledge to Robotic Control},
  author={Zitkovich, Brianna and Joshi, Nikhil and Lohan, Aviral and others},
  booktitle={Proceedings of The 7th Conference on Robot Learning},
  year={2023}
}

@article{roboflamingo,
  title={Vision-Language Foundation Models as Effective Robot Imitators},
  author={Li, Xinghang and Liu, Minghuan and Zhang, Hanbo and Yu, Cunjun and Xu, Jie and Wu, Hongtao and Cheang, Chilam and Jing, Ya and Zhang, Weinan and Liu, Huaping and Kong, Tao},
  journal={arXiv preprint arXiv:2311.01378},
  year={2023}
}

@article{openvla,
  title={{OpenVLA}: An Open-Source Vision-Language-Action Model},
  author={Kim, Moo Jin and Pertsch, Karl and Karamcheti, Siddharth and Xiao, Ted and Balakrishna, Ashwin and Nair, Suraj and Rafailov, Rafael and others},
  journal={arXiv preprint arXiv:2406.09246},
  year={2024}
}

@article{pi0,
  title={A Vision-Language-Action Flow Model for General Robot Control},
  author={Black, Kevin and Brown, Noah and Darpinian, James and others},
  journal={arXiv preprint arXiv:2410.24164},
  year={2024}
}

@article{pi05,
  title={$\pi_{0.5}$: A Vision-Language-Action Model with Open-World Generalization},
  author={{Physical Intelligence} and Black, Kevin and Brown, Noah and others},
  journal={arXiv preprint arXiv:2504.16054},
  year={2025}
}

@article{gr00t-n1,
  title={{GR00T N1}: An Open Foundation Model for Generalist Humanoid Robots},
  author={{NVIDIA} and Bjorck, Johan and Casta\~neda, Fernando and Cherniadev, Nikita and Da, Xingye and Ding, Runyu and Fan, Linxi and others},
  journal={arXiv preprint arXiv:2503.14734},
  year={2025}
}

@misc{gr00t-n15,
  author       = {{NVIDIA Research}},
  title        = {{GR00T N1.5}: An Improved Open Foundation Model for Generalist Humanoid Robots},
  year         = {2025},
  month        = jun,
  howpublished = {Technical blog / model release},
  url          = {https://research.nvidia.com/labs/gear/gr00t-n1_5/},
  note         = {Accessed: 2026-05-05}
}

@misc{gr00t-n16,
  author       = {{NVIDIA Research}},
  title        = {{GR00T N1.6}: An Improved Open Foundation Model for Generalist Humanoid Robots},
  year         = {2025},
  month        = dec,
  howpublished = {Technical blog / model release},
  url          = {https://research.nvidia.com/labs/gear/gr00t-n1_6/},
  note         = {Accessed: 2026-05-05}
}

@misc{gr00t-n17,
  author       = {{NVIDIA}},
  title        = {{NVIDIA Isaac GR00T N1.7}: Open Reasoning VLA Model for Humanoid Robots},
  year         = {2026},
  month        = apr,
  howpublished = {Hugging Face community article},
  url          = {https://huggingface.co/blog/nvidia/gr00t-n1-7},
  note         = {Accessed: 2026-05-05}
}

@inproceedings{tothenoiseandback,
  title={To the Noise and Back: Diffusion for Shared Autonomy},
  author={Yoneda, Takuma and Sun, Luzhe and Yang, Ge and Stadie, Bradly C. and Walter, Matthew R.},
  booktitle={Robotics: Science and Systems},
  year={2023}
}

@article{bid,
  title={Bidirectional Decoding: Improving Action Chunking via Closed-Loop Resampling},
  author={Liu, Yuejiang and Hamid, Jubayer Ibn and Xie, Annie and Lee, Yoonho and Du, Maximilian and Finn, Chelsea},
  journal={arXiv preprint arXiv:2408.17355},
  year={2024}
}

@article{rtc,
  title={Real-Time Execution of Action Chunking Flow Policies},
  author={Black, Kevin and Galliker, Manuel Y. and Levine, Sergey},
  journal={arXiv preprint arXiv:2506.07339},
  year={2025}
}

@article{dtp,
  title={{DTP}: A Simple yet Effective Distracting Token Pruning Framework for Vision-Language Action Models},
  author={Li, Chenyang and Liu, Jieyuan and Li, Bin and Gao, Bo and Yuan, Yilin and He, Yangfan and Li, Yuchen and Tang, Jingqun},
  journal={arXiv preprint arXiv:2601.16065},
  year={2026}
}

@inproceedings{damo,
  title={{DAMO}: Decoding by Accumulating Activations Momentum for Mitigating Hallucinations in Vision-Language Models},
  author={Wang, Kaishen and Gu, Hengrui and Gao, Meijun and Zhou, Kaixiong},
  booktitle={The Thirteenth International Conference on Learning Representations},
  year={2025}
}

@article{tang2025dcla,
  title={Mitigating Hallucinations via Inter-Layer Consistency Aggregation in Large Vision-Language Models},
  author={Tang, Kai and You, Jinhao and Ge, Xiuqi and Li, Hanze and Guo, Yichen and Huang, Xiande},
  journal={arXiv preprint arXiv:2505.12343},
  year={2025}
}

@article{zhuang2025vasparse,
  title={{VASparse}: Towards Efficient Visual Hallucination Mitigation for Large Vision-Language Model via Visual-Aware Sparsification},
  author={Zhuang, Xianwei and Zhu, Zhihong and Xie, Yuxin and Liang, Liming and Zou, Yuexian},
  journal={arXiv preprint arXiv:2501.06553},
  year={2025}
}

@article{ki,
  title={Knowledge Insulating Vision-Language-Action Models: Train Fast, Run Fast, Generalize Better},
  author={Driess, Danny and Springenberg, Jost Tobias and Ichter, Brian and Yu, Lili and Li-Bell, Adrian and Pertsch, Karl and Ren, Allen Z. and Walke, Homer and Vuong, Quan and Shi, Lucy Xiaoyang and Levine, Sergey},
  journal={arXiv preprint arXiv:2505.23705},
  year={2025}
}

@article{reconvla,
  title={{ReconVLA}: Reconstructive Vision-Language-Action Model as Effective Robot Perceiver},
  author={Song, Wenxuan and Zhou, Ziyang and Zhao, Han and Chen, Jiayi and Ding, Pengxiang and Yan, Haodong and Huang, Yuxin and Tang, Feilong and Wang, Donglin and Li, Haoang},
  journal={arXiv preprint arXiv:2508.10333},
  year={2025}
}

@article{cgvd,
  title={Overcoming Visual Clutter in Vision Language Action Models via Concept-Gated Visual Distillation},
  author={Song, Sangmim and Kodagoda, Sarath and Carmichael, Marc and Thiyagarajan, Karthick},
  journal={arXiv preprint arXiv:2603.10340},
  year={2026}
}

@article{pointvla,
  title={Point What You Mean: Visually Grounded Instruction Policy},
  author={Yu, Hang and Zhao, Juntu and Liu, Yufeng and Li, Kaiyu and Ma, Cheng and Zhang, Di and Hu, Yingdong and Chen, Guang and Xie, Junyuan and Guo, Junliang and Zhao, Junqiao and Gao, Yang},
  journal={arXiv preprint arXiv:2512.18933},
  year={2025}
}

@article{demodiffusion,
  title={Demodiffusion: One-shot human imitation using pre-trained diffusion policy},
  author={Park, Sungjae and Bharadhwaj, Homanga and Tulsiani, Shubham},
  journal={arXiv preprint arXiv:2506.20668},
  year={2025}
}

@inproceedings{hajl,
  title={Human-agent joint learning for efficient robot manipulation skill acquisition},
  author={Luo, Shengcheng and Peng, Quanquan and Lv, Jun and Hong, Kaiwen and Driggs--Campbell, Katherine Rose and Lu, Cewu and Li, Yong--Lu},
  booktitle={2025 IEEE International Conference on Robotics and Automation (ICRA)},
  pages={1370--1377},
  year={2025},
  organization={IEEE}
}

@article{vls,
  title={{VLS}: Steering Pretrained Robot Policies via Vision-Language Models},
  author={Liu, Shuo and Singh, Ishneet Sukhvinder and Xu, Yiqing and Duan, Jiafei and Krishna, Ranjay},
  journal={arXiv preprint arXiv:2602.03973},
  year={2026}
}

@article{vlareasoner,
  title={{VLA-Reasoner}: Empowering Vision-Language-Action Models with Reasoning via Online Monte Carlo Tree Search},
  author={Guo, Wenkai and Lu, Guanxing and Deng, Haoyuan and Wu, Zhenyu and Tang, Yansong and Wang, Ziwei},
  journal={arXiv preprint arXiv:2509.22643},
  year={2025}
}

@article{taco,
  title={Steering Vision-Language-Action Models as Anti-Exploration: A Test-Time Scaling Approach},
  author={Yang, Siyuan and Zhang, Yang and He, Haoran and Pan, Ling and Li, Xiu and Bai, Chenjia and Li, Xuelong},
  journal={arXiv preprint arXiv:2512.02834},
  year={2025}
}

@inproceedings{dynaguide,
  title={DynaGuide: Steering Diffusion Policies with Active Dynamic Guidance},
  author={Du, Maximilian and Song, Shuran},
  booktitle={Advances in Neural Information Processing Systems},
  year={2025}
}

@article{liberopro,
  title={{LIBERO-PRO}: Towards Robust and Fair Evaluation of Vision-Language-Action Models Beyond Memorization},
  author={Zhou, Xueyang and Xu, Yangming and Tie, Guiyao and Chen, Yongchao and Zhang, Guowen and Chu, Duanfeng and Zhou, Pan and Sun, Lichao},
  journal={arXiv preprint arXiv:2510.03827},
  year={2025}
}

@article{simplerenv,
  title={Evaluating real-world robot manipulation policies in simulation},
  author={Li, Xuanlin and Hsu, Kyle and Gu, Jiayuan and Pertsch, Karl and Mees, Oier and Walke, Homer Rich and Fu, Chuyuan and Lunawat, Ishikaa and Sieh, Isabel and Kirmani, Sean and others},
  journal={arXiv preprint arXiv:2405.05941},
  year={2024}
}

@inproceedings{pope,
  title     = {Evaluating Object Hallucination in Large Vision-Language Models},
  author    = {Li, Yifan and Du, Yifan and Zhou, Kun and Wang, Jinpeng and Zhao, Wayne Xin and Wen, Ji-Rong},
  booktitle = {Proceedings of the 2023 Conference on Empirical Methods in Natural Language Processing},
  pages     = {292--305},
  year      = {2023},
  address   = {Singapore},
  publisher = {Association for Computational Linguistics},
  doi       = {10.18653/v1/2023.emnlp-main.20},
  url       = {https://aclanthology.org/2023.emnlp-main.20}
}

@inproceedings{vcd,
  title     = {Mitigating Object Hallucinations in Large Vision-Language Models through Visual Contrastive Decoding},
  author    = {Leng, Sicong and Zhang, Hang and Chen, Guanzheng and Li, Xin and Lu, Shijian and Miao, Chunyan and Bing, Lidong},
  booktitle = {Proceedings of the IEEE/CVF Conference on Computer Vision and Pattern Recognition (CVPR)},
  pages     = {13872--13882},
  year      = {2024},
  month     = {June}
}

@inproceedings{opera,
  title     = {{OPERA}: Alleviating Hallucination in Multi-Modal Large Language Models via Over-Trust Penalty and Retrospection-Allocation},
  author    = {Huang, Qidong and Dong, Xiaoyi and Zhang, Pan and Wang, Bin and He, Conghui and Wang, Jiaqi and Lin, Dahua and Zhang, Weiming and Yu, Nenghai},
  booktitle = {Proceedings of the IEEE/CVF Conference on Computer Vision and Pattern Recognition (CVPR)},
  pages     = {13418--13427},
  year      = {2024},
  month     = {June}
}

@inproceedings{m3id,
  title     = {Multi-Modal Hallucination Control by Visual Information Grounding},
  author    = {Favero, Alessandro and Zancato, Luca and Trager, Matthew and Choudhary, Siddharth and Perera, Pramuditha and Achille, Alessandro and Swaminathan, Ashwin and Soatto, Stefano},
  booktitle = {Proceedings of the IEEE/CVF Conference on Computer Vision and Pattern Recognition (CVPR)},
  pages     = {14303--14312},
  year      = {2024},
  month     = {June}
}

@article{gaze_regularized_vla,
  title   = {Gaze-Regularized Vision-Language-Action Models for Robotic Manipulation},
  author  = {Pani, Anupam and Yang, Yanchao},
  journal = {arXiv preprint arXiv:2603.23202},
  year    = {2026},
  url     = {https://arxiv.org/abs/2603.23202}
}

@article{univla,
  title={Univla: Learning to act anywhere with task-centric latent actions},
  author={Bu, Qingwen and Yang, Yanting and Cai, Jisong and Gao, Shenyuan and Ren, Guanghui and Yao, Maoqing and Luo, Ping and Li, Hongyang},
  journal={arXiv preprint arXiv:2505.06111},
  year={2025}
}

@inproceedings{aim,
  title={Aim: Adaptive inference of multi-modal llms via token merging and pruning},
  author={Zhong, Yiwu and Liu, Zhuoming and Li, Yin and Wang, Liwei},
  booktitle={Proceedings of the IEEE/CVF International Conference on Computer Vision},
  pages={20180--20192},
  year={2025}
}

@article{llava,
  title={Visual instruction tuning},
  author={Liu, Haotian and Li, Chunyuan and Wu, Qingyang and Lee, Yong Jae},
  journal={Advances in neural information processing systems},
  volume={36},
  pages={34892--34916},
  year={2023}
}

@article{qwen3,
  title={Qwen3-vl technical report},
  author={Bai, Shuai and Cai, Yuxuan and Chen, Ruizhe and Chen, Keqin and Chen, Xionghui and Cheng, Zesen and Deng, Lianghao and Ding, Wei and Gao, Chang and Ge, Chunjiang and others},
  journal={arXiv preprint arXiv:2511.21631},
  year={2025}
}

@article{paligemma,
  title={Paligemma: A versatile 3b vlm for transfer},
  author={Beyer, Lucas and Steiner, Andreas and Pinto, Andr{\'e} Susano and Kolesnikov, Alexander and Wang, Xiao and Salz, Daniel and Neumann, Maxim and Alabdulmohsin, Ibrahim and Tschannen, Michael and Bugliarello, Emanuele and others},
  journal={arXiv preprint arXiv:2407.07726},
  year={2024}
}

@inproceedings{prismatic,
  title={Prismatic vlms: Investigating the design space of visually-conditioned language models},
  author={Karamcheti, Siddharth and Nair, Suraj and Balakrishna, Ashwin and Liang, Percy and Kollar, Thomas and Sadigh, Dorsa},
  booktitle={Forty-first International Conference on Machine Learning},
  year={2024}
}

@article{eaglevlm,
  title={Eagle 2: Building post-training data strategies from scratch for frontier vision-language models},
  author={Li, Zhiqi and Chen, Guo and Liu, Shilong and Wang, Shihao and VS, Vibashan and Ji, Yishen and Lan, Shiyi and Zhang, Hao and Zhao, Yilin and Radhakrishnan, Subhashree and others},
  journal={arXiv preprint arXiv:2501.14818},
  year={2025}
}

@inproceedings{molmo,
  title={Molmo and pixmo: Open weights and open data for state-of-the-art vision-language models},
  author={Deitke, Matt and Clark, Christopher and Lee, Sangho and Tripathi, Rohun and Yang, Yue and Park, Jae Sung and Salehi, Mohammadreza and Muennighoff, Niklas and Lo, Kyle and Soldaini, Luca and others},
  booktitle={Proceedings of the Computer Vision and Pattern Recognition Conference},
  pages={91--104},
  year={2025}
}

@article{ema,
  title={Forecasting seasonals and trends by exponentially weighted moving averages},
  author={Holt, Charles C},
  journal={International journal of forecasting},
  volume={20},
  number={1},
  pages={5--10},
  year={2004},
  publisher={Elsevier}
}

@article{obeyed_vla,
  title={Clutter-Resistant Vision-Language-Action Models through Object-Centric and Geometry Grounding},
  author={Vo, Khoa and Hanyu, Taisei and Ikebe, Yuki and Pham, Trong Thang and Chung, Nhat and Vu, Minh Nhat and Minh, Duy Nguyen Ho and Nguyen, Anh and Gunderman, Anthony and Rainwater, Chase and others},
  journal={arXiv preprint arXiv:2512.22519},
  year={2025}
}

@article{tag,
  title={{TAG}: Target-Agnostic Guidance for Stable Object-Centric Inference in Vision-Language-Action Models},
  author={Zhou, Jiaying and Zhan, Zhihao and Zhai, Ruifeng and Lyu, Qinhan and Liu, Hao and Wang, Keze and Lin, Liang and Wang, Guangrun},
  journal={arXiv preprint arXiv:2603.24584},
  year={2026}
}

@article{ivra,
  title={{IVRA}: Improving Visual-Token Relations for Robot Action Policy with Training-Free Hint-Based Guidance},
  author={Park, Jongwoo and Ranasinghe, Kanchana and Jang, Jinhyeok and Mata, Cristina and Jang, Yoo Sung and Ryoo, Michael S},
  journal={arXiv preprint arXiv:2601.16207},
  year={2026}
}

@article{vlasurvey_token,
  title={A survey on vision-language-action models: An action tokenization perspective},
  author={Zhong, Yifan and Bai, Fengshuo and Cai, Shaofei and Huang, Xuchuan and Chen, Zhang and Zhang, Xiaowei and Wang, Yuanfei and Guo, Shaoyang and Guan, Tianrui and Lui, Ka Nam and others},
  journal={arXiv preprint arXiv:2507.01925},
  year={2025}
}

@article{coavla,
  title={CoA-VLA: Improving vision-language-action models via visual-textual chain-of-affordance},
  author={Li, Jinming and Zhu, Yichen and Tang, Zhibin and Wen, Junjie and Zhu, Minjie and Liu, Xiaoyu and Li, Chengmeng and Cheng, Ran and Peng, Yaxin and Peng, Yan and others},
  journal={arXiv preprint arXiv:2412.20451},
  year={2024}
}

@inproceedings{cogvla,
  title={CogVLA: Cognition-Aligned Vision-Language-Action Models via Instruction-Driven Routing \& Sparsification},
  author={Li, Wei and Zhang, Renshan and Shao, Rui and He, Jie and Nie, Liqiang},
  booktitle={The Thirty-ninth Annual Conference on Neural Information Processing Systems},
  year={2025}
}

@article{vlasurvey,
  title={A Survey on Vision--Language--Action Models for Embodied AI},
  author={Ma, Yueen and Song, Zixing and Zhuang, Yuzheng and Hao, Jianye and King, Irwin},
  journal={IEEE Transactions on Neural Networks and Learning Systems},
  year={2026},
  publisher={IEEE}
}

@article{vlasurvey_realworld_app,
  title={Vision-language-action models for robotics: A review towards real-world applications},
  author={Kawaharazuka, Kento and Oh, Jihoon and Yamada, Jun and Posner, Ingmar and Zhu, Yuke},
  journal={IEEE Access},
  year={2025},
  publisher={IEEE}
}

@article{vlasurvey_concept,
  title={Vision-language-action models: Concepts, progress, applications and challenges},
  author={Sapkota, Ranjan and Cao, Yang and Roumeliotis, Konstantinos I and Karkee, Manoj},
  journal={arXiv preprint arXiv:2505.04769},
  year={2025}
}

@article{vlm4vla,
  title={VLM4VLA: Revisiting Vision-Language-Models in Vision-Language-Action Models},
  author={Zhang, Jianke and Chen, Xiaoyu and Wang, Qiuyue and Li, Mingsheng and Guo, Yanjiang and Hu, Yucheng and Zhang, Jiajun and Bai, Shuai and Lin, Junyang and Chen, Jianyu},
  journal={arXiv preprint arXiv:2601.03309},
  year={2026}
}

@article{gpt2,
  title={Language models are unsupervised multitask learners},
  author={Radford, Alec and Wu, Jeffrey and Child, Rewon and Luan, David and Amodei, Dario and Sutskever, Ilya and others},
  journal={OpenAI blog},
  volume={1},
  number={8},
  pages={9},
  year={2019}
}

@inproceedings{resnet,
  title={Deep residual learning for image recognition},
  author={He, Kaiming and Zhang, Xiangyu and Ren, Shaoqing and Sun, Jian},
  booktitle={Proceedings of the IEEE conference on computer vision and pattern recognition},
  pages={770--778},
  year={2016}
}

@article{transformer,
  title={Attention is all you need},
  author={Vaswani, Ashish and Shazeer, Noam and Parmar, Niki and Uszkoreit, Jakob and Jones, Llion and Gomez, Aidan N and Kaiser, {\L}ukasz and Polosukhin, Illia},
  journal={Advances in neural information processing systems},
  volume={30},
  year={2017}
}

@article{haelm,
  title={Evaluation and analysis of hallucination in large vision-language models},
  author={Wang, Junyang and Zhou, Yiyang and Xu, Guohai and Shi, Pengcheng and Zhao, Chenlin and Xu, Haiyang and Ye, Qinghao and Yan, Ming and Zhang, Ji and Zhu, Jihua and others},
  journal={arXiv preprint arXiv:2308.15126},
  year={2023}
}

@inproceedings{hallusionbench,
  title={Hallusionbench: an advanced diagnostic suite for entangled language hallucination and visual illusion in large vision-language models},
  author={Guan, Tianrui and Liu, Fuxiao and Wu, Xiyang and Xian, Ruiqi and Li, Zongxia and Liu, Xiaoyu and Wang, Xijun and Chen, Lichang and Huang, Furong and Yacoob, Yaser and others},
  booktitle={Proceedings of the IEEE/CVF conference on computer vision and pattern recognition},
  pages={14375--14385},
  year={2024}
}

@article{amber,
  title={Amber: An llm-free multi-dimensional benchmark for mllms hallucination evaluation},
  author={Wang, Junyang and Wang, Yuhang and Xu, Guohai and Zhang, Jing and Gu, Yukai and Jia, Haitao and Wang, Jiaqi and Xu, Haiyang and Yan, Ming and Zhang, Ji and others},
  journal={arXiv preprint arXiv:2311.07397},
  year={2023}
}

@article{mllm_hallucination_survey,
  title={A survey of multimodal hallucination evaluation and detection},
  author={Chen, Zhiyuan and Min, Yuecong and Zhang, Jie and Yan, Bei and Wang, Jiahao and Wang, Xiaozhen and Shan, Shiguang},
  journal={International Journal of Computer Vision},
  volume={134},
  number={3},
  pages={131},
  year={2026},
  publisher={Springer}
}

@article{action_hallucination,
  title={Action Hallucination in Generative Visual-Language-Action Models},
  author={Soh, Harold and Lim, Eugene},
  journal={arXiv preprint arXiv:2602.06339},
  year={2026}
}

@article{simonyan2013saliency,
  title={Deep Inside Convolutional Networks: Visualising Image Classification Models and Saliency Maps},
  author={Simonyan, Karen and Vedaldi, Andrea and Zisserman, Andrew},
  journal={arXiv preprint arXiv:1312.6034},
  year={2013}
}

\newpage
\appendix
\section{Internal Mechanism of \algo}
\label{app:mechanism}

We have shown in Section~\ref{sec:experiments} that \algo improves VLA performance at inference time.
This section reveals the internal effect of \algo.

\subsection{Experiment Setup}
\label{app:experiment_setup}

All experiments in this section are conducted with the $\pi_{0.5}$ VLA model on the LIBERO-PRO benchmark.
We record two quantities at layer $k$.
First, we record the hidden state of the action expert after the layer, denoted by $h^k$.
Second, we record the action-to-image attention map, denoted by $A^k$.
For masking diagnostics, we compare the original unmasked visual input with masked visual inputs obtained by replacing pixels inside a selected mask region with the mean RGB value of the image, while keeping all other inputs unchanged.
This masking operation removes local visual evidence.
Figure~\ref{fig:app_object_awareness} summarizes the complete diagnostic protocol and the main empirical observations used in this section.

\subsection{Object-awareness of Action Hidden State}
\label{app:object_awareness_definition}

A VLA policy ultimately predicts actions from action representations rather than from attention visualizations.
Therefore, the diagnostic question we ask is not only whether the model visually attends to the target object, but whether the task-relevant object evidence actually affects the hidden state used for action generation.
To measure this effect, we define an awareness score based on object~masking.

For a mask region $m$, let $h^k_{\varnothing}$ be the layer-$k$ hidden state under the original visual input, and let $h^k_m$ be the hidden state under the corresponding masked input.
We define the layer-wise awareness induced by mask $m$ as
\begin{equation}
\mathcal{A}^k(m)
=
\frac{
\left\|h^k_{\varnothing}-h^k_m\right\|_2
}{
\left\|h^k_{\varnothing}\right\|_2
}.
\label{eq:layer_awareness}
\end{equation}
When $m$ is the task-relevant object mask $m_o$, the resulting quantity $\mathcal{A}^k(m_o)$ is the object awareness of the layer.
A larger value means that masking the task object produces a larger relative change in the action hidden state, indicating stronger sensitivity to task-relevant object evidence.

In Section~\ref{sec:introduction}, we introduce object awareness in its simplest form, using the task-object mask:
$\Vert h^k_{(\mathrm{m})} - h^k_{(\mathrm{o})}\Vert / \Vert h^k_{(\mathrm{o})}\Vert$.
We now generalize the same idea into a mask-dependent awareness function to conduct controlled experiments.
Under this definition, $\mathcal{A}^k(m_o)$ recovers the object-awareness score introduced in Section~\ref{sec:introduction}.

To verify that object awareness is not merely measuring generic sensitivity to image corruption, we construct two controls: the same number of pixels randomly scattered across the image ($m_r$), and a rectangular region of similar area randomly sampled from the background ($m_b$).
We define control awareness as
\begin{equation}
\mathcal{A}^k_{\mathrm{ctrl}}
=
\max
\left\{\mathcal{A}^k(m_b),\mathcal{A}^k(m_r)\right\}.
\label{eq:control_awareness}
\end{equation}
To summarize the masking comparison, we use two statistics: object-awareness gain and object-selectivity ratio. They quantify two complementary effects: whether \algo increases sensitivity to the task-relevant object compared with native inference, and whether the measured sensitivity is stronger for the task-relevant object than for same-area control regions.

The object-awareness gain compares \algo with native inference:
\begin{equation}
R^k_{\mathrm{gain}}
=
\frac{
\mathcal{A}^k_{\text{\algo}}(m_o)
}{
\mathcal{A}^k_{\mathrm{native}}(m_o)
}.
\label{eq:awareness_gain}
\end{equation}
The object-selectivity ratio compares object masking with same-area control masking under the same inference method $u\in\{\mathrm{native},\text{\algo}\}$:
\begin{equation}
R^k_{\mathrm{select}}(u)
=
\frac{
\mathcal{A}^k_u(m_o)
}{
\mathcal{A}^k_{u,\mathrm{ctrl}}
}.
\label{eq:awareness_selectivity}
\end{equation}
Thus, $R^k_{\mathrm{gain}}>1$ indicates that \algo increases object awareness relative to native inference, while $R^k_{\mathrm{select}}>1$ indicates that the measured awareness is stronger for the task object than for same-area control regions.

\begin{figure*}[t]
    \centering
    \includegraphics[width=\textwidth]{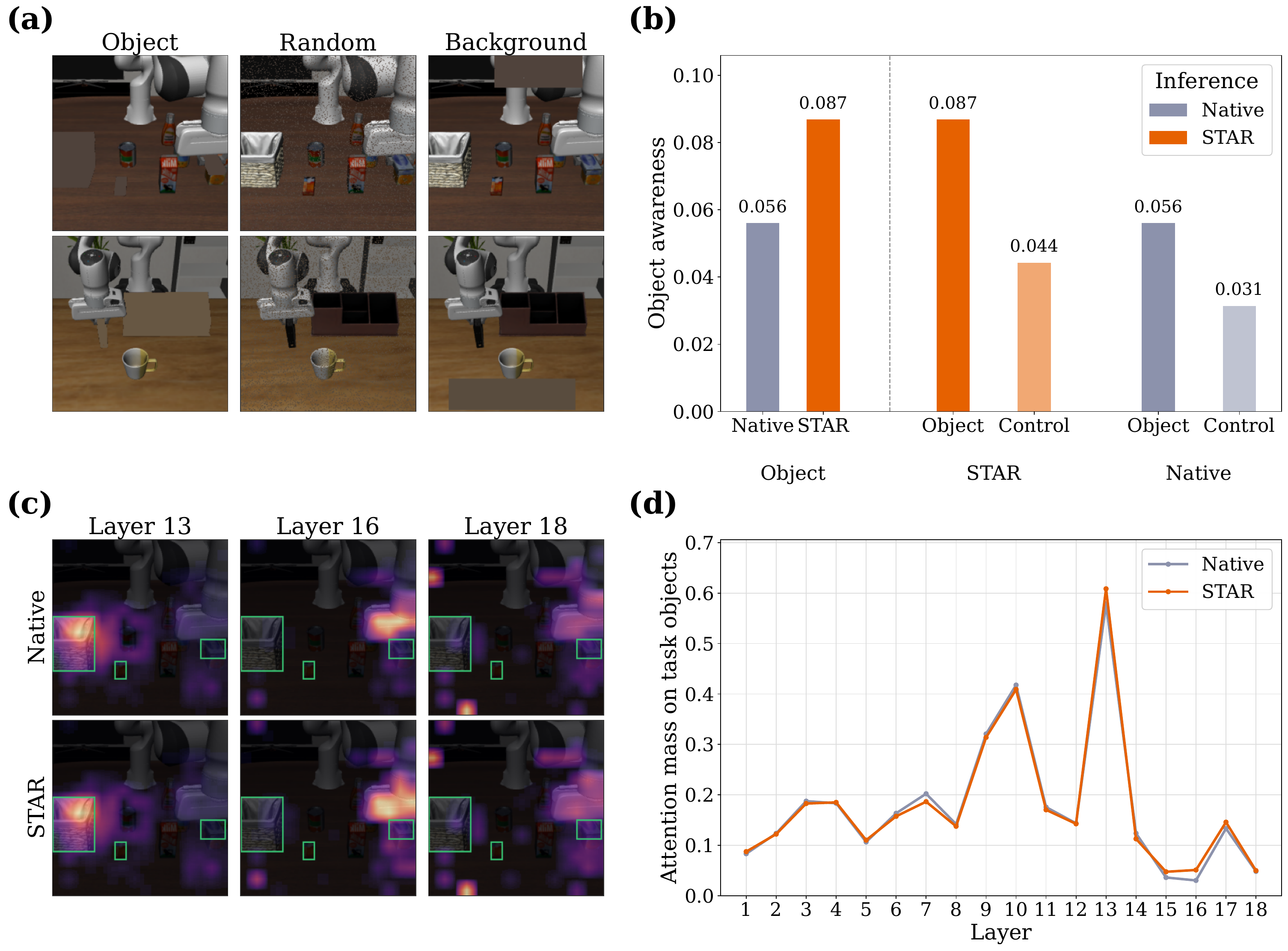}
    \caption{
    Object-awareness and attention-map diagnostics for native inference and \algo.
    (a) Visualization of the three masks, including the task-specific object mask $m_o$, randomly scattered pixels $m_r$ with the same pixel count, and a randomly sampled rectangular background region $m_b$ of similar area.
    (b) Comparison of awareness scores for native inference and \algo under object masking and control masking at layer 18.
    (c) Attention-map visualization at layers 13, 16, and 18, showing that native inference and \algo produce similar attention patterns.
    (d) Attention mass on the task-object region across layers.
    Native inference and \algo exhibit close attention-mass curves, indicating that attention-based object focus does not clearly reveal the hidden-state intervention effect.
    }
    \label{fig:app_object_awareness}
\end{figure*}

\label{app:awareness_effective}
Using the above metric, we compare native inference and \algo on the paired original and masked inputs.
Figure~\ref{fig:app_object_awareness}(a) visualizes the three mask types used in the diagnostic, and
Figure~\ref{fig:app_object_awareness}(b) reports the corresponding average awareness scores at layer 18 under native inference and \algo.
Under the experimental setting with the $\pi_{0.5}$ VLA model on the LIBERO-PRO benchmark, \algo increases object awareness from $0.056$ to $0.087$, giving $R_{\mathrm{gain}}=1.554$.

The selectivity statistics further show that the measured awareness is task-object specific.
For native inference, $R_{\mathrm{select}}(\mathrm{native})=1.806$.
For \algo, $R_{\mathrm{select}}(\text{\algo})=1.977$.
Therefore, the increased awareness under \algo is not merely a response to arbitrary masking. It is stronger when the task-relevant object region is removed than when same-area control regions are removed.

These results show that object awareness is an effective diagnostic for the hidden-state intervention.
It clearly captures that \algo makes late action representations more sensitive to task-relevant object evidence.
As shown later in Section~\ref{app:attention_ineffective}, this object-awareness signal is much clearer than the corresponding attention-map change.

\subsection{Gradient-based Saliency Map}
\label{app:gradient_saliency}

The masking analysis above measures whether the action hidden state is sensitive to the task object. We further ask a complementary, action-grounded question: does the final predicted action rely more strongly on pixels belonging to that object? To answer this question, we use input-gradient saliency~\citep{simonyan2013saliency}, which identifies the image evidence that locally influences the action output.

For a predicted action chunk $a_{1:K}$, we first average the actions in the chunk,
\begin{equation}
    u = \frac{1}{K}\sum_{k=1}^{K} a_k,
\end{equation}
and construct a scalar action score by projecting $u$ onto its stop-gradient direction:
\begin{equation}
    s =
    u^\top
    \operatorname{sg}\!\left(
        \frac{u}{\lVert u\rVert_2+\epsilon}
    \right),
\end{equation}
where $\operatorname{sg}(\cdot)$ denotes stop-gradient. Holding the action direction fixed during differentiation gives a scalar whose input gradient measures how each image region supports the action selected by the policy, without allowing the reference direction itself to change during backpropagation.

Let $I$ be the input image and let $g(p)$ denote the magnitude of $\partial s/\partial I$ at pixel $p$, aggregated over the color channels. We spatially normalize this evidence and define the target-object gradient score as
\begin{equation}
    \mathcal{S}_{\mathrm{grad}}
    =
    \frac{
        \sum_{p\in\Omega_o} g(p)
    }{
        \sum_{p\in\Omega_{\mathrm{img}}} g(p)+\epsilon
    },
\end{equation}
where $\Omega_o$ is the task-object region and $\Omega_{\mathrm{img}}$ contains all image pixels. Thus, a larger $\mathcal{S}_{\mathrm{grad}}$ means that a greater fraction of the action-relevant visual evidence lies on the key object.

\begin{figure*}[b]
    \centering
    \includegraphics[width=\textwidth]{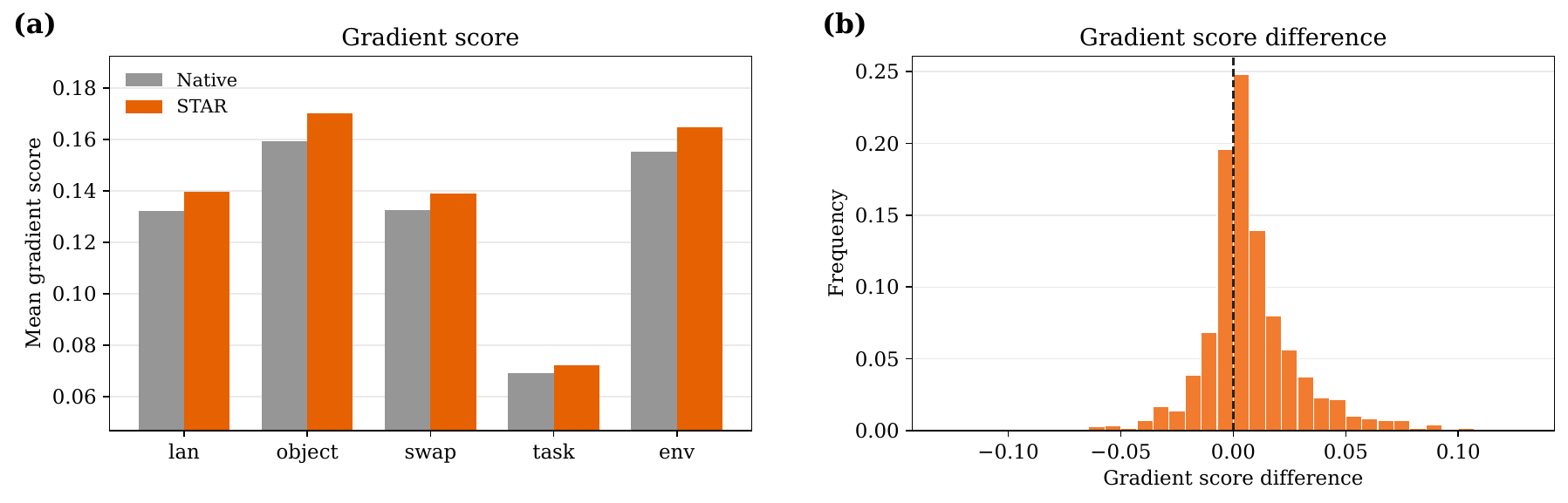}
    \caption{
    \textbf{Action-gradient scores on LIBERO-PRO.}
    (a) Mean normalized input-gradient evidence contained in the task-object region for native inference and \algo across the five LIBERO-PRO evaluation splits. Higher values indicate that the predicted action depends more strongly on the key object. \algo consistently increases the score on all five splits.
    (b) Distribution of paired score differences, computed as \algo minus native inference. Most of the mass lies to the right of zero, showing that the increase is consistent across evaluation samples rather than being driven only by split-level averages.
    }
    \label{fig:app_grad}
\end{figure*}

Figure~\ref{fig:app_grad}(a) shows that \algo achieves a higher target-object gradient score than native inference on every LIBERO-PRO split. The improvement is therefore not restricted to a particular perturbation type: it appears under language, object-attribute, spatial-configuration, task, and visual-context variation. The paired differences in Figure~\ref{fig:app_grad}(b) are also predominantly positive, indicating that the aggregate gains reflect a broadly consistent sample-level trend.
These results are consistent with the final action depending more strongly on task-object pixels under STAR.

\begin{figure*}[t]
    \centering
    \includegraphics[width=0.97\textwidth]{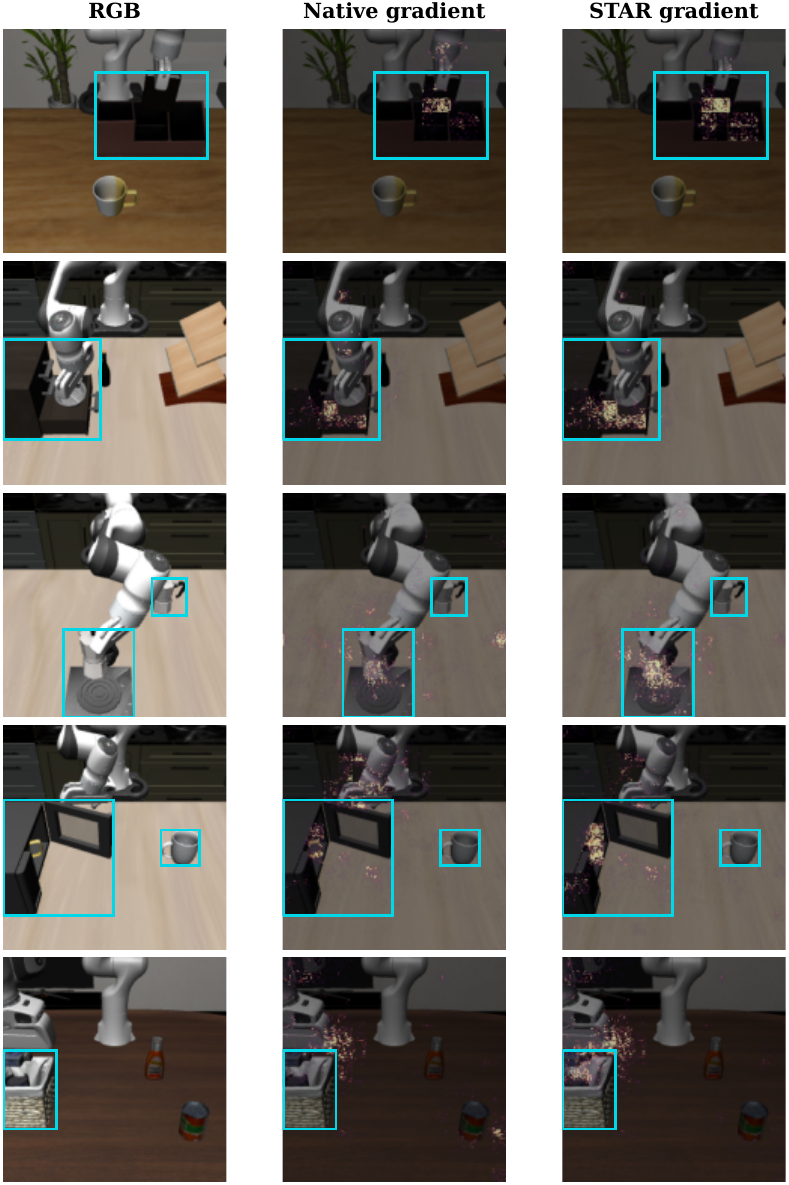}
    \caption{
    \textbf{Qualitative action-gradient saliency maps.}
    The left column shows the RGB observations with task-relevant objects marked by cyan boxes. The middle and right columns show saliency under native inference and \algo, respectively. Warmer pixels exert a stronger local influence on the predicted action. Compared with native inference, \algo concentrates more action-relevant evidence within the marked object regions across diverse manipulation scenes.
    }
    \label{fig:app_grad_vis}
\end{figure*}

\clearpage

The qualitative maps in Figure~\ref{fig:app_grad_vis} reveal the corresponding spatial effect. Under native inference, the action gradients are often weaker on the task object or dispersed over the robot and background. With \algo, stronger evidence is concentrated inside the cyan task-object regions, including articulated fixtures, manipulated objects, and interaction targets. Importantly, this result complements the attention analysis in Section~\ref{app:attention_ineffective}: \algo need not visibly reallocate attention mass to make the action more object-grounded. Instead, by carrying object-aware residual-update directions into the late action representations, it increases the local sensitivity of the final action to key-object pixels. Together, the masking and gradient diagnostics support the proposed mechanism: \algo preserves task-relevant visual evidence in the hidden-state update stream and makes that evidence more consequential for action generation.

\subsection{Attention Maps Do Not Reveal the Intervention Effect}
\label{app:attention_ineffective}

We next test whether attention maps can reveal the effect of the hidden-state intervention.
We compare native inference and \algo using the same input and record the action-to-image attention maps at late action-expert layers.
Figure~\ref{fig:app_object_awareness}(c) visualizes representative attention maps from native inference and \algo at layers 13, 16, and 18.
Although \algo modifies the hidden-state stream, the corresponding attention maps remain visually similar across these layers.

To provide a more quantitative comparison, we compute the attention mass assigned to the task-object region.
Let $\Omega_o$ denote the image-token set covered by the task-object mask, $\Omega_{\mathrm{img}}$ denote all image tokens, and $A^k$ denote the attention map of layer $k$.
For method $u\in\{\mathrm{native},\text{\algo}\}$, the layer-wise object attention mass is
\begin{equation}
\mathcal{M}^k_u
=
\frac{
\sum_{p\in\Omega_o} A^k_u(p)
}{
\sum_{p\in\Omega_{\mathrm{img}}} A^k_u(p)
}.
\label{eq:attention_mass}
\end{equation}
Figure~\ref{fig:app_object_awareness}(d) reports $\mathcal{M}^k_u$ across layers for native inference and \algo.
These results show that the attention-based measure of object focus remains close to that of native inference and does not reliably expose the hidden-state intervention.

\subsection{Diagnostics of STAR's Internal Mechanism}
\label{app:diagnostic_conclusion}

The diagnostics above suggest that the internal effect of \algo is primarily reflected in the action hidden state and its dependence on visual evidence rather than in a visible reallocation of action-to-image attention.
Object masking shows that \algo increases the sensitivity of late action representations to task-relevant object evidence, while same-area control masks induce smaller responses. Gradient saliency further shows that a larger fraction of the visual evidence influencing the final action is concentrated on the task object.
In contrast, the attention maps and attention-mass curves remain close to native inference.
Thus, \algo preserves task-object information in late action representations and makes it more consequential for action generation without necessarily producing a salient change in attention visualizations.

\section{Hyperparameter Settings}
\label{app:hyperparameters}

Table~\ref{tab:hyperparameters} reports the hyperparameters used for each backbone and benchmark. The coefficient notation follows Section~\ref{sec:method}, with $w_\omega$ denoting the weight for orientation change.

\begin{table}[h]
\centering
\footnotesize
\setlength{\tabcolsep}{3.2pt}
\renewcommand{\arraystretch}{1.15}
\caption{
\textbf{Hyperparameter settings for each backbone and benchmark.}
}
\label{tab:hyperparameters}
\begin{tabular}{@{}l l c c c c c c c@{}}
\toprule
\textbf{Backbone} &
\textbf{Benchmark} &
$\boldsymbol{\beta}_{\min}$ &
$\boldsymbol{\beta}_{\max}$ &
$\boldsymbol{w}_c$ &
$\boldsymbol{w}_v$ &
$\boldsymbol{w}_j$ &
$\boldsymbol{w}_\omega$ &
$\boldsymbol{\kappa}$ \\
\midrule
$\pi_{0.5}$ & LIBERO-PRO & 0.25 & 0.55 & 0.25 & 1.0 & $-0.125$ & $-0.25$ & 2.30 \\
GR00T-N1.5 & LIBERO-PRO & 0.20 & 0.50 & 1.0 & 1.0 & $-0.375$ & $-2.0$ & 3.75 \\
GR00T-N1.5 & SimplerEnv & 0.20 & 0.50 & 1.0 & 1.0 & $-0.375$ & $-2.0$ & 4.00 \\
OpenVLA & SimplerEnv & 0.25 & 0.55 & 1.0 & 1.0 & $-0.25$ & $-0.75$ & 4.75 \\
\bottomrule
\end{tabular}
\end{table}

For the $\pi_{0.5}$ success-rate evaluation on LIBERO-PRO, we apply STAR to the final six action-expert layers. The environment runs at 10 Hz, and the policy executes five actions before replanning.

\end{document}